\documentclass[a4paper,fleqn]{cas-dc}

\usepackage[authoryear,longnamesfirst]{natbib}
\usepackage{hyperref} 

\def\tsc#1{\csdef{#1}{\textsc{\lowercase{#1}}\xspace}}
\tsc{WGM}
\tsc{QE}
\tsc{EP}
\tsc{PMS}
\tsc{BEC}
\tsc{DE}

\begin{document}
\let\WriteBookmarks\relax
\def\floatpagepagefraction{1}
\def\textpagefraction{.001}
\shorttitle{Uniﬁed Modality-Speciﬁc Representation}
\shortauthors{Zamzmi et~al.}

\title [mode = title]{UMS-Rep: Unified Modality-Specific Representation for Efficient Medical Image Analysis}

\author[1]{Ghada Zamzmi}[type=editor,
                        ]


\address[1]{Lister Hill National Center for Biomedical Communications, National Institutes of Health, Bethesda, MD, USA}

\author[1]{Sivaramakrishnan Rajaraman}[type=editor]

\author[1]{Sameer Antani}[%
   type=editor,
   ]







\begin{abstract}
Medical image analysis typically includes several tasks such as enhancement, segmentation, and classification. Traditionally, these tasks are implemented using separate deep learning models for separate tasks, which is not efficient because it involves unnecessary training repetitions, demands greater computational resources, and requires a relatively large amount of labeled data. In this paper, we propose a multi-task training approach for medical image analysis, where individual tasks are fine-tuned simultaneously through relevant knowledge transfer using a unified modality-specific feature representation (UMS-Rep). We explore different fine-tuning strategies to demonstrate the impact of the strategy on the performance of target medical image tasks. We experiment with different visual tasks (e.g., image denoising, segmentation, and classification) to highlight the advantages offered with our approach for two imaging modalities, chest X-ray and Doppler echocardiography. Our results demonstrate that the proposed approach reduces the overall demand for computational resources and improves target task generalization and performance. Specifically, the proposed approach improves accuracy (up to $\sim$ 9\% $\uparrow$) and decreases computational time (up to $\sim$ 86\% $\downarrow$) as compared to the baseline approach. Further, our results prove that the performance of target tasks in medical images is highly influenced by the utilized fine-tuning strategy.
\end{abstract}









\maketitle

\section{Introduction}
Deep learning has significantly advanced the frontiers of visual task analysis, with applications in a large number of domains including automated driving, robotics, biometrics, and biomedical imaging, to name a few. With breakthrough advances in medical image analysis, there has been a surge in research interest in health applications and biomedical research. Some common visual tasks across various medical imaging modalities include (i) region of interest (ROI) detection (e.g., \cite{zhou2017deep,litjens2017survey}), (ii) segmentation (e.g., \cite{litjens2017survey,mahapatra2019progressive,gulati2020application,christodoulidis2016multi}), (iii) registration (e.g., \cite{balakrishnan2018unsupervised}), and (iv) classification (e.g., \cite{litjens2017survey, mahapatra2019progressive,gulati2020application}). 

While these approaches have achieved promising performance, they tend to use individual pre-trained models (e.g., VGG16 and ResNet with ImageNet weights) for individual tasks. Training and deploying multiple individual models with high computational complexity for multiple tasks is feasible, but is constrained by limited computational resources and repetitive sequential training. These factors can significantly impact future advances and pose challenges in deploying applications with a large number of models in real-world clinical settings. Additionally, using as many models as there are tasks limits the potential offered by deep learning by constraining it from transferring knowledge from the source domain to a target domain or task.

These challenges can be alleviated through multi-task learning (MTL), a machine learning concept that transfers the knowledge from a shared source to multiple target tasks (\cite{thung2018brief}), thereby improving target tasks generalizability and aiding overall performance. The target tasks can be homogeneous with similar annotations or heterogeneous with diverse annotations. For example, two target tasks, such as semantic and instance segmentation, that use similar annotations can be jointly learned via a shared encoder and task-specific decoders. Joint learning strategy involves learning a joint loss function that takes into consideration the inter-task relationship. Other fine-tuning strategies include independent and alternating strategies \cite{thung2018brief,dong2015multi,zhou2019models}. We further discuss these strategies in Section 2.3.

MTL techniques have been widely used for natural images but few researches \cite{moeskops2016deep,kisilev2016medical} have been carried out to apply these techniques on medical images. Inspired by the success of MTL with natural images, we propose an MTL-based training approach in medical image analysis, where individual target tasks are learned simultaneously through relevant knowledge transfer across a shared medical modality-specific feature representation. This approach enables reduced requirements for computational resources and enhances target tasks generalization.

The main contributions of this paper are as follows:
\begin{itemize}
\item We propose an approach that uses a shared source, called unified modality-specific representation (UMS-Rep), to simultaneously fine-tune target medical image tasks with diverse and similar annotations.
\item We show that the shared source (UMS-Rep) can be constructed on a specific medical imaging modality using any learning techniques including unsupervised (ex nihilo without annotations) or supervised (with limited annotations). Also, we show that UMS-Rep can simultaneously learn pre-processing tasks such as image denoising and enhancement.
\item We explore three fine-tuning strategies, namely independent, alternating, and joint, to investigate their impact on the performance of target tasks in medical images.
\item We define derivable tasks, which can be any task derived from the learned target tasks. An example of a derivable task includes visual interpretation of machine learning decisions, derived from a classification target task.
\end{itemize}

Our empirical evaluations show that the proposed sequence of training each part of the network, first the shared UMS-Rep with pre-processing tasks, second the task-specific heads (target tasks) with the suitable fine-tuning strategy, and finally derivable tasks, is more efficient for medical image analysis. Particularly, our evaluations on medical image benchmarks demonstrate the superiority and efficiency of our approach, in terms of performance, resources, and computation, as compared to the baseline approach of training as many models as there are tasks.

The rest of this paper is organized as follows. Section 2 discusses different types of target tasks and fine-tuning strategies. Section 3 presents the medical image datasets used in this paper and describes our training approach for medical image analysis. Then, we present the experimental results and a comparison with the baseline approach in Section 4. Finally, we discuss the results and conclude the paper in Section 5.

\section{Background}
MTL methods have attracted attention as they show remarkable success in improving the generalizability of target tasks (\cite{thung2018brief}). These target tasks can be divided based on data annotations into homogeneous tasks (similar annotation) and heterogeneous tasks (diverse annotation). Next, we present homogeneous and heterogeneous target tasks and discuss different strategies for fine-tuning. 

\subsection{Homogeneous Target Tasks}
MTL-based methods have been widely used to jointly learn target tasks with similar annotations. For example, \cite{zhang2014facial} proposed a MTL-based method to jointly learn the following target tasks from a face image: pose, gender, wear glasses, and smiling. These target tasks have global image labels (presence/absence of a smile, a glass, etc.) as the ground truth (GT) annotations. The inter-task relationship or correlation was learned using a joint loss function as presented in (\cite{zhang2014facial}). Another method is proposed by \cite{elhoseiny2015convolutional} to jointly learn the category and pose of objects. Similar to \cite{zhang2014facial}, both target tasks have image-level label annotations (i.e., category label and pose label) and the relationship between these tasks is learned using a joint loss function. Other methods for learning target tasks with homogeneous annotations are presented in \cite{kendall2018multi,moeskops2016deep,chen2018multi,bai2019self}. 

\subsection{Heterogeneous Target Tasks}
\cite{teichmann2018multinet} proposed a method for modeling three heterogeneous tasks in natural images, namely street classification, vehicle detection, and road segmentation, with diverse annotations (global, box coordinates, and pixel-level labels). Each task is trained individually, and the final loss is given as the sum of the losses for segmentation, detection, and classification. YOLO network (\cite{redmon2016you}) is another example of a method that learns tasks with diverse annotations. Specifically, YOLO uses a head for the bounding box (coordinates) and another head for classification (image label). YOLO uses a loss function with a localization loss term for the bounding box task and classification loss for the classification task. 

Unlike natural images, medical images are not only limited in the number of annotations, but they also exhibit different visual characteristics (e.g., color homogeneity, texture subtlety, shape variation, and highly similar appearance). An approach has been attempted by \cite{zhou2019models} for learning heterogeneous medical image tasks. They built a set of models, called Genesis, and used them as the starting point to learn classification and segmentation tasks. Both tasks are fine-tuned independently using two individual loss functions.

  \begin{figure}[!t]
        \centering
        \includegraphics[height=0.28\textwidth]{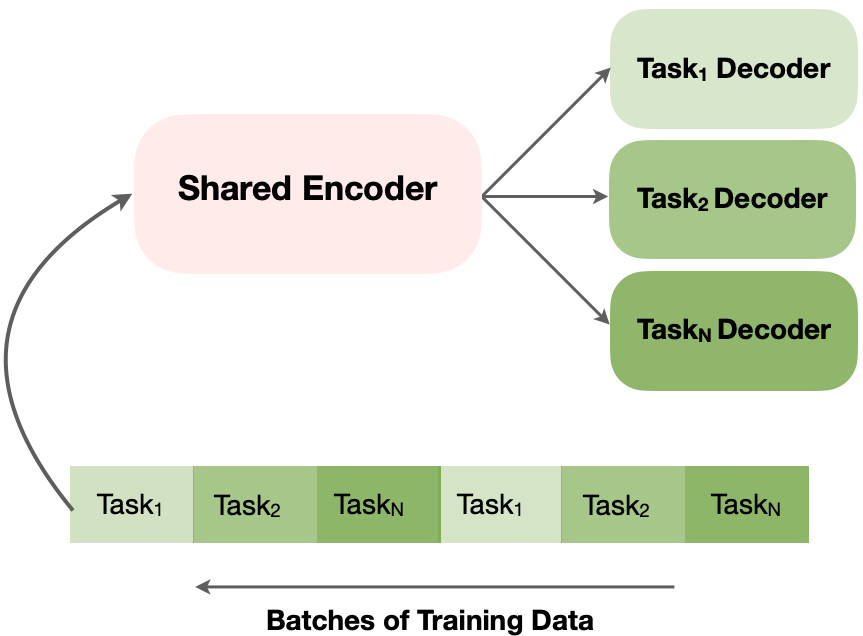}
        \caption{The fine-tuning approach in alternating strategy}
        \label{fig:my_label}
    \end{figure}

\subsection{Fine-tuning Strategies}
We group the fine-tuning strategies for the target tasks into: joint fine-tuning, alternating fine-tuning, and independent fine-tuning. 

Joint fine-tuning strategy involves optimizing the entire network to minimize a single loss function that accumulates the losses of all target tasks. This strategy allows learning relationships between target tasks. The weights ($w_i$) of losses can be either weighted or unweighted. In case of unweighted summation, the loss functions for all tasks receive equal weights. Otherwise, the optimal weights for tasks’ losses can be hand-tuned or learned through extensive empirical experiments. Other approaches for weighting losses include homoscedastic uncertainty which was introduced by \cite{kendall2018multi} and cross-task consistency as proposed by \cite{zou2018df}.


Instead of optimizing a single joint function, the alternating strategy requires each target task to have a separate loss function; i.e., $N$ loss functions for $N$ tasks (\cite{dong2015multi}). The loss function for each task has two terms, the first term ($\Theta_{src}$) represents the parameters for the shared source and the second term represents the task-specific parameters ($\Theta_{T_N}$) for $N$ tasks. The optimization process is performed by alternately optimizing the loss of a specific target task using a task-specific batch followed by optimizing the loss of another task using its specific batch, and so on (\cite{dong2015multi}). Figure 1 shows the optimization procedure for alternating fine-tuning strategy, which involves fine-tuning a specific task for a fixed number of batches before switching to the batches of the next task (\cite{dong2015multi}). In each switch, we update the task-specific loss with its terms ($\Theta_{src}$ and $\Theta_{T_i}$). The ratio of task-specific batches can be fixed for all tasks, determined based on other factors such as performance, importance, and dataset size, or calculated using specific methods such as the method proposed in (\cite{luong2015multi}). By alternating the learning among tasks and updating the weights ($\Theta_{src}$ and $\Theta_{T_{i=1:N}}$), this strategy allows to learn the similar latent representations across the target tasks.

Finally, the independent strategy involves fine-tuning target tasks independently while freezing $\Theta_{src}$. This strategy allows all target tasks to share a common representation followed by task-specific layers. Each task-specific head is fine-tuned using its own loss and optimizer. In other words, we freeze the shared representation parameters ($\Theta_{src}$) and use task-specific optimizers to minimize task-specific losses on task-specific data.

In summary, joint fine-tuning strategy optimizes the entire network (shared source and task-specific heads) by minimizing a single joint function that accumulates the losses of all target tasks. The alternating strategy optimizes a specific-task by minimizing its loss for a fixed number of batches before switching to the next task. In each switch, both the source and task-specific parameters are updated. In the independent fine-tuning strategy, we freeze the parameters of the shared representation and independently optimize task-specific heads by minimizing their losses. It has been reported (\cite{standley2020tasks}) that independent strategy leads to better performance when used for fine-tuning competing (or different) tasks while joint strategy leads to better performance with cooperating (or similar) tasks. 
This is attributed to the fact that competing tasks may lead to the transfer of irrelevant information when learned jointly (i.e., negative transfer between unrelated tasks). Further, the gradients may interfere, which makes the optimization landscape of multiple summed losses more difficult.

\begin{figure}[!t]
\begin{center}
\includegraphics[height=0.60\textwidth]{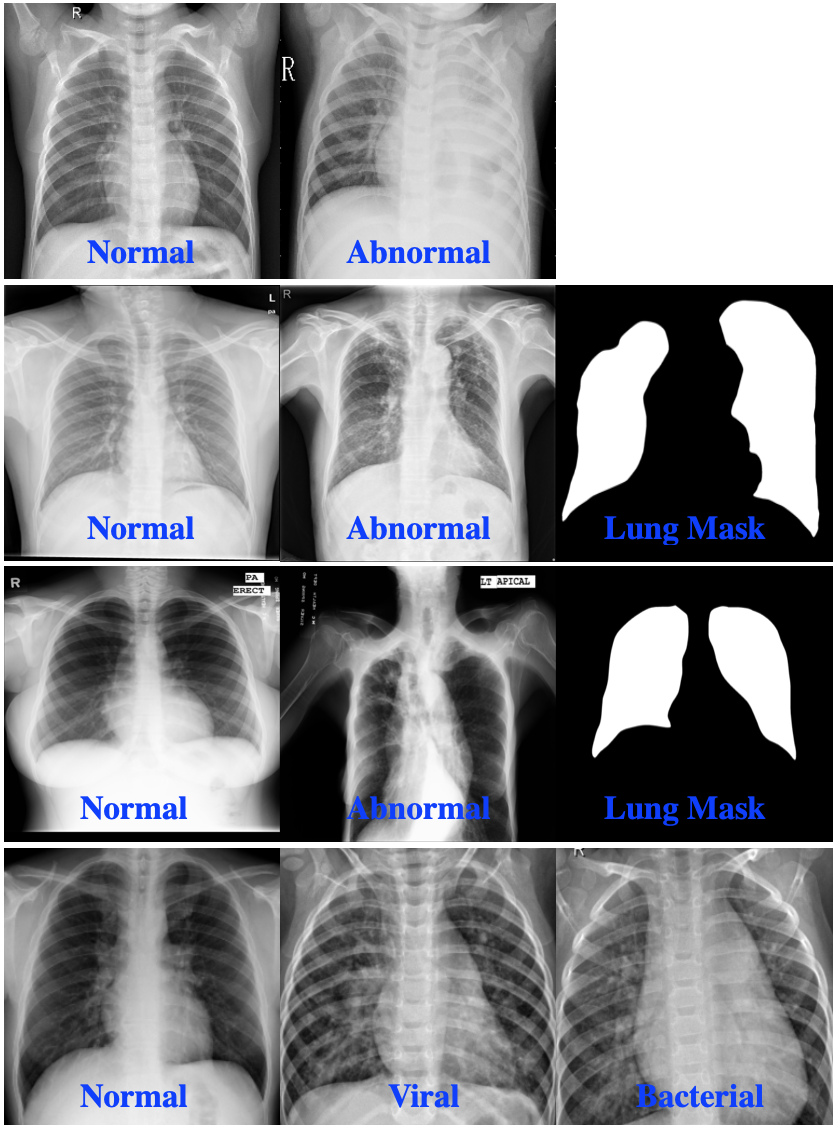}
\caption{Samples from CXR datasets and their annotations; 1$^{st}$ row: RSNA, 2$^{nd}$ row: Shenzhen TB CXR, 3${rd}$ row: Montgomery TB CXR, and 4$^{th}$ row: Pediatric pneumonia CXR.}
\end{center}
\end{figure}

\begin{figure}[!t]
\begin{center}
\includegraphics[height=0.30\textwidth]{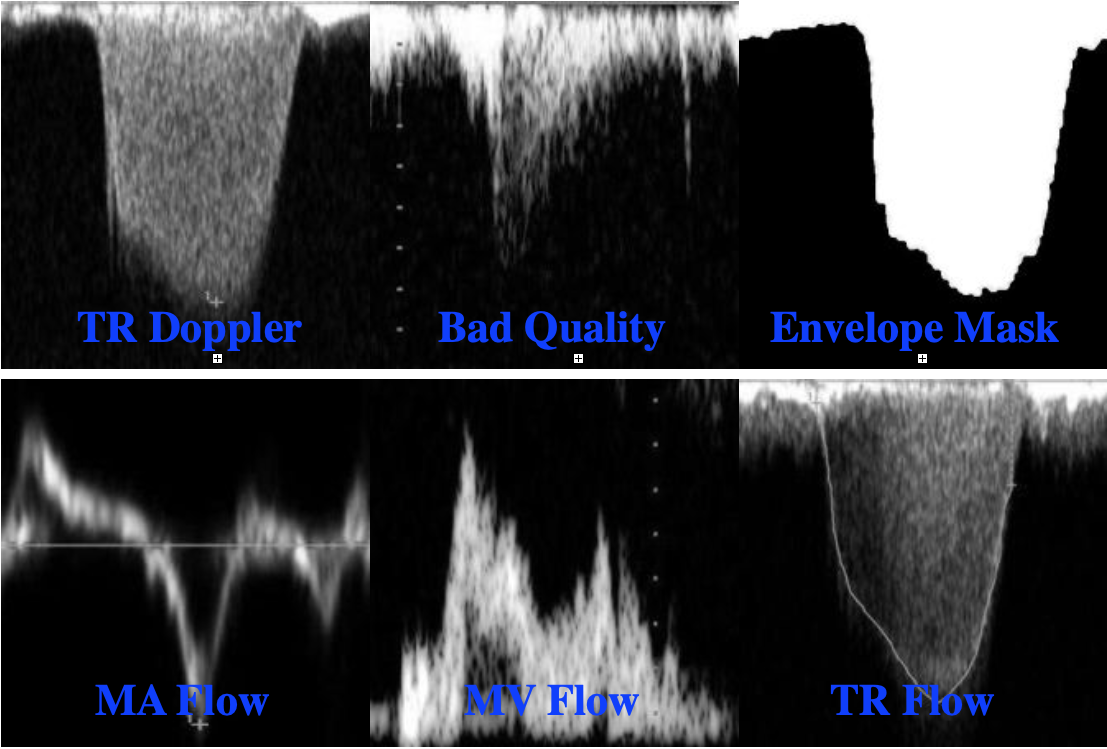}
\caption{Samples from echo Doppler datasets and their annotations; 1$^{st}$ row: Doppler tasks and 2$^{nd}$ row: Doppler flows.}
\end{center}
\end{figure}

\begin{table*}[!t] 
  \begin{tabular}
      {|l|l|l|l|l|} 

      \hline \textbf{Modality} & \textbf{Dataset} & \textbf{Total Images} & \textbf{GT Labels} & \textbf{Resolution} \\ 
      \hline
       & RSNA & 26,684 & Normal  & 1024 x 1024 
\\
       &  &  & Abnormal & 
       \\
      &  &  &  &  
      \\
      
       & Shenzhen & 662 & Normal (336) & 3000 x 3000 
\\
       &  &  & Abnormal (326) & 
       \\
       &  &  & Lung mask (566) &  \\
       &  &  &  &    \\
       
       CXR & Montgomery & 138 & Normal (80) & 4020x 4892 
       \\
       &  &  & Abnormal (58) &  
       \\
        &  &  & Lung mask (138) & 
        \\
       &  &  &  &  
       \\
       
        & Pediatric  & Train: 5,232 & Bacterial (2,538) & -- 
        \\
       & Pneumonia &  & Viral (1,345 ) & 
       \\
        &  &  & Normal (1,349)& 
        \\
       & & Test: 624  &  Bacterial (242) & 
       \\
       & &  &  Viral (148) &  
       \\
       & &  &  Normal (234) &  
       \\
       &  &  &  &  
       \\
       \hline
       &  &  &  &  
       \\
         & Doppler  & 2444 & Doppler Flow (2444) & -- 
         \\
       & Tasks &  & Image Quality (814) &  
       \\
        &  &  & Envelope mask (2444) & 
        \\
       &  &  &  &  
       \\
       Doppler &  &  &  &  
       \\
       Echo &  &  &  &  
       \\
        & Doppler  &  & MV Flow (855) &  
        \\
       & Flows &  & MA Flow(490) &  
       \\
        &  &  & TR Flow (1099) & 
        \\
       &  &  &  &  
       \\
      
        &  &  &  &  
        \\
       
      \hline
  \end{tabular}
  \caption{Summary of CXR and Doppler echo Datasets.}
\end{table*}

\section{Materials and Methods}
\subsection{Medical Image Datasets}
Two medical imaging modalities are used to evaluate the proposed approach described in Section 3.2. These modalities are chest X-ray (CXR) and Doppler echocardiography (Doppler echo). Table 1 provides a summary of the CXR and Doppler echo datasets used in this work.  

\subsubsection{CXR Collections}
We used four publicly available CXR collections in our evaluation of the proposed approach. These collections are the Radiological Society of North America (RSNA) \cite{shih2019augmenting}, the Shenzhen TB \cite{jaeger2014two}, the Montgomery TB \cite{jaeger2014two}, and the Pediatric pneumonia dataset \cite{kermany2018identifying}. 

The RSNA dataset includes 26,684 normal and abnormal frontal CXRs that are provided as DICOM images with $1024 \times 1024$ spatial resolution. In addition to the labels, GT disease bounding boxes are made available for CXRs containing pneumonia-related opacities. The Shenzhen dataset contains 662 CXR images (336 abnormal and 326 normal image-level labels) and GT binary masks for the lungs (pixel-level labels). The size of the images in this collection varies, but it is approximately $3000 \times 3000$ pixels. The Montgomery dataset has 138 posterior-anterior CXRs of which 80 CXRs are normal and 58 are abnormal with TB manifestations. The images have $4020 \times 4892$ pixel resolution. The abnormal class for Montgomery dataset includes a wide range of abnormalities, including tuberculosis-related manifestations, effusions, and miliary patterns. Finally, the Pediatric pneumonia dataset was collected from 624 patients and labeled as described in \cite{kermany2018identifying}. The train set contains a total of 5,232 chest X-ray images, including 3,883 labeled as pneumonia (2,538 bacterial and 1,345 viral) and 1,349 labeled as normal. The test set contains 234 normal images and 390 pneumonia images (242 bacterial and 148 viral). Figure 2 shows labeled image examples from the aforementioned CXR datasets.

The images of all datasets were resized to $256 \times 256$ using bicubic interpolation (OpenCV built-in function). We also performed mean normalization to ensure that the images have a similar distribution. As it is well known, data normalization can speed up convergence while training the network.

\subsubsection{Doppler Echo}
We used a private dataset of 2444 images showing continuous wave and pulsed wave Doppler flows collected from 100 patients who were referred for echocardiographic examination in the Clinical Center at the National Institutes of Health (NIH). The use of these de-identified images was approved by the NIH Ethics Review Board (IRB\#18-NHLBI-00686). The Doppler traces of the mitral valve flow (MV), mitral annular flow (MA), and tricuspid regurgitation flow (TR) were acquired using different commercial echocardiography systems including Philips iE33, GE Vivid95, and GE Vivid9. Each Doppler image has a flow type label (TR, MV, or MA) and a segmentation mask provided by an expert technician, which separates the spectral envelope from the background.  Besides, the expert technician assessed the quality of a subset of images (814 out of 2444) as low- or good-quality. All GT annotations (global labels and binary masks) provided by the expert technician were further verified by an expert cardiologist. 
Figure 3 shows labeled image examples from the Doppler echo dataset.

All the images were resized to $256 \times 256$ using bicubic interpolation (OpenCV built-in function). We then performed mean normalization.

\subsection{Proposed Approach}
The traditional approach for medical image analysis (e.g., \cite{christodoulidis2016multi,zhou2017deep}) is shown in Figure 4. This approach involves training $N$ individual (encoder-decoder) models for $N$ individual target tasks in isolation. Our proposed approach of simultaneously training UMS-Rep with pre-processing tasks and sharing the trained UMS-Rep among target tasks is presented in Figure 5.

\subsubsection{Notations and Definitions}
As shown in Figure 5, UMS-Rep learns the visual features ($X_S$) of a source input space or specific imaging modality ($S$). The learned feature representation is then shared to simultaneously learn $N$ target tasks $T_{1:N} = \{T_1,T_2,..,T_N\}$. The source modality is defined as $S$ = \{$X_S, P_S(X_S)$\}, where $X_S$ represents the feature space and $P_S(X_S)$ represents the probability distribution of the input modality. 

\begin{figure}[!t]
\begin{center}
\includegraphics[width=0.85\linewidth]{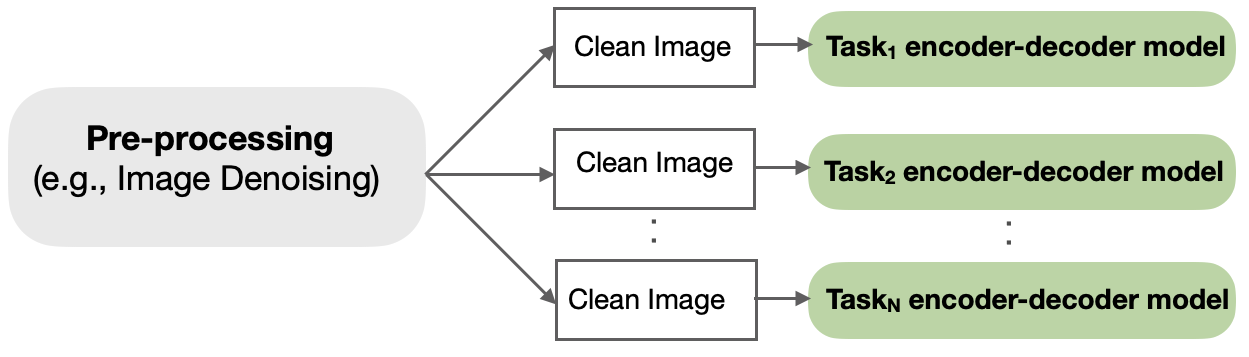}
\end{center}
\caption{The traditional sequence of training in medical image analysis involves using individual models for individual tasks. For example, denoising medical images prior to the analysis requires using a separate image denoising model to generate clean images followed by using these images as input to individual models. }
\label{fig:long}
\label{fig:onecol}
\end{figure}

\begin{figure}[!t]
\begin{center}
\includegraphics[width=0.90\linewidth]{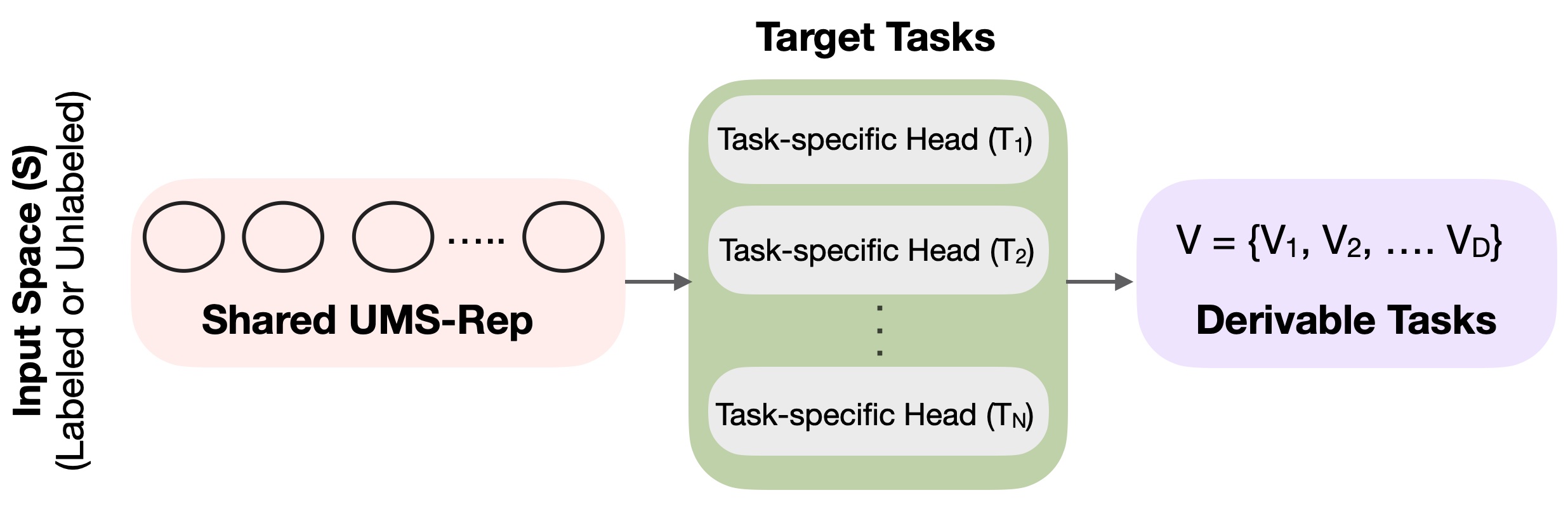}

\end{center}
\caption{Our proposed sequence of simultaneously training UMS-Rep with pre-processing tasks (pink box). UMS-Rep is then shared among target tasks with heterogeneous or homogeneous annotations. $T_{1:N} = \{T_1,.., T_N \}$ and $V_{1:D} = \{V_1,.., V_D\}$ represent target tasks (green box) and derivable tasks (purple task), respectively.}
\label{fig:long}
\label{fig:onecol}
\end{figure}

UMS-Rep improves the learning and generalization of the predictive functions of up to $N$ target tasks ($f_{1:N}(.)=\{f_1(.), f_2(.), .., f_N(.)\}$) by transferring the shared knowledge from $S$ to $N$ to generate task-specific predictions $Y_{1:N}= \{Y_1,..,Y_N\}$. Since medical image analysis tasks that use a common input or modality share common low-level features, a shared modality-specific representation (source) can be trained once and used to simultaneously fine-tune target tasks with diverse or similar GT labels. For example, the features extracted from a specific imaging modality ($S$) by an autoencoder using unsupervised learning ($Y_{GT_S} = \{\}$) can be simultaneously shared by two target tasks $T_1$ and $T_2$ with different label spaces, where $P_{S}(X_S) \sim P_1(X_1)$ and $P_{S}(X_S) \sim P_2(X_2)$. A new task-specific head (or decoder) can be added on-the-fly if this constraint is satisfied. 

In Figure 5, we also define D derivable tasks $V_{1:D} = \{V_1, V_2, ..., V_D\}$. These tasks are the ones that use the information learned by a single or a combination of task-specific heads. An example of a derivable task is the interpretation or visual explanation generated from classification task prediction. Such a task is crucial for augmenting medical decision making (\cite{rajaraman2018visualization}). Another derivable task is automated decision recommendation, which can be generated by combining the outputs of different task-specific heads. A new derivable task can also be added on-the-fly with minimal effort as long as it uses information of previously learned task-specific heads to produce the output.

Our proposed sequence of learning each part, first the shared UMS-Rep with pre-processing tasks, second the task-specific heads, and finally the derivable tasks, allows more efficient analysis of medical images. 

\subsubsection{UMS-Rep Construction}
Given that the proposed approach holds the promise for simultaneously reducing computation/resources and enhancing the generalization of $N$ target tasks, a question that arises is: what learning technique should be used to construct the shared UMS-Rep? The answer depends on the input space and the tasks at hand. For example, given a dataset with different label spaces for different tasks, UMS-Rep can be constructed using the task with the largest number of labeled data (supervised learning). Instead of using the labels of a specific task to construct the shared UMS-Rep, an autoencoder can be optimized for a given source modality ($S$) and used as a shared representation for simultaneously learning multiple target tasks (unsupervised learning). 

In this paper, we experimented with both unsupervised and supervised learning techniques to construct the shared UMS-Rep. Note that other techniques (e.g., semi-supervised, residual learning) can be easily adopted to construct UMS-Rep. In both cases, we used shallow customized architectures for UMS-Rep; however, the state-of-the-arts architectures (e.g., VGG and UNet) can be used instead. While constructing UMS-Rep, a pre-processing task (i.e., noise reduction) is also learned. Using the proposed approach for simultaneously learning pre-processing tasks and solving visual tasks can prevent unnecessary repetitions, reduce computations, and enhance generalizability.

\subsubsection{Target Tasks Layers \& Fine-tuning Strategies}
Prior to fine-tuning the target tasks, we appended task-specific layers (or decoders) to the constructed UMS-Rep. To create the task-specific layers for the segmentation task, we append a symmetrical decoder to the shared UMS-Rep. In case of classification, we append a global average pooling (GAP), fully connected (FC), dropout (D), and Softmax (SM) layers to the shared UMS-Rep. Then, we fine-tune the task-specific layers using three strategies, namely alternating, independent, and joint, to investigate the impact of the fine-tuning strategy on the performance of the target tasks. 

As discussed in Section 2.3, joint fine-tuning involves learning a single loss that combines the sum of unweighted or weighted losses for all target tasks. There are different approaches (e.g., \cite{zhang2014facial,elhoseiny2015convolutional,moeskops2016deep,kendall2018multi}) for weighting losses. In this paper, we learned, through empirical experiments, the optimal weights for tasks' losses. Learning a single loss function that weighs losses of the target tasks allows learning the commonality and differences between target tasks. On the other hand, alternating strategy (see Figure 1) learns target tasks by alternating the learning among tasks and updating the weights. This strategy allows to simultaneously 1) transfer common features from the shared source to task-specific layers and 2) learn similar latent representations across the target tasks. We determine the ratio of task-specific batches as described in \cite{luong2015multi}. After determining the ratio, we train a specific task $i$ for $n_i$ batches and switch to the next task $j$, which is trained for $n_j$ batches. Finally, we use the independent fine-tuning strategy, where we freeze the weights of the shared UMS-Rep and fine-tune each target task independently using its own loss and optimizer. 

The target tasks can be optimized using different loss functions. In this paper, we optimize the target classiﬁcation tasks to minimize the categorical cross-entropy (CCE) loss. For the segmentation tasks, we minimize a combination of binary cross-entropy (BCE) and Dice losses as follows:

\begin{equation}
    L_{n} = w_1 L_{BCE_n} + w_2 L_{DSC_n} 
\end{equation}

where $Loss_{BCE_n}$ is the binary cross-entropy, $Loss_{DSC_n}$ is the Dice loss, $n$ denotes the batch number, and $w_1=w_2=0.5$. The losses are computed for each mini-batch, and the final loss for the entire batch is determined by the mean of loss across all the mini-batches. $Loss_{BCE_n}$ and $Loss_{DSC_n}$ are expressed as follows:

\begin{equation}
    L_{BCE_n} = - [t_n log(y_n) + (1-t_n) log(1-y_n)]
\end{equation}

and,
\begin{equation}
    L_{DSC_n} = 1 - \frac{2 \sum{ t_n \cdot y_n}}{\sum t_n + \sum y_n}
\end{equation}

where $t$ is the target and $y$ is the output from the final layer. We empirically found that the combination of $L_{BCE_n}$ and $L_{DSC_n}$ losses (equation 1) improved the optimization due to the interplay between the global and local feature extraction capabilities of these loss functions.

\section{Experiments and Results}

We evaluated our approach on two medical imaging modalities: CXR and echo Doppler. To construct UMS-Rep, we experimented with two learning paradigms: unsupervised (CXR) and supervised (echo Doppler). Then, the constructed UMS-Reps, for each modality, are appended with task-specific layers corresponding to different target tasks. These layers are fine-tuned using independent, alternating, or joint strategies. In our work, the input domains of all target tasks have similar distributions to the input domain of UMS-Reps.

To demonstrate the power of the proposed approach, we compare the performance with the traditional baseline approach presented in Figure 4, in terms of accuracy and computations. The baseline approach for medical image analysis requires a separate model (encoder-decoder) to learn each pre-processing task in addition to $N$ encoders and $N$ decoders to learn $N$ target tasks. On the other hand, the proposed approach requires only a single encoder and $N$ lightweight task-specific decoders for $N$ target tasks, resulting in significant savings in time and the number of training parameters.

Our experiments provide empirical answers to the following questions:
\begin{itemize}
    \item What advantages does the proposed sequence of training offer in comparison to the baseline approach in terms of performance and computational efficiency?
    \item What fine-tuning strategy should be used for the target medical image tasks? 
\end{itemize}

To report the performance of target tasks, we used F-score, Matthews Correlation Coefficient (MCC), and the accuracy. We also used the intersection over union (IoU) metric to report the performance of the segmentation task. To demonstrate the computational efficiency, we reported the computational times and the training parameters for all models. 

We trained all models using a Windows system with the following configuration: (1) Intel Xeon CPU E3-1275 v6 3.80 GHz and (2) NVIDIA GeForce GTX 1050 Ti. Keras DL framework with Tensorflow was used for model training and evaluation. For hyperparameter optimization, Talos\footnote{https://github.com/autonomio/talos} library with Keras was used. The code of this work is made publicly available at \href{https://github.com/GhadaZ/UMSRep}{UMS-Rep Github Page}.

\subsection{CXR Imaging Modality}
Although the high-level (i.e., task-specific) features of CXR target tasks are relatively different, CXR images have similar low-level features (e.g., edges and blobs). Hence, UMS-Rep can be constructed to learn CXR low-level features and then shared among different CXR tasks.

\begin{figure*}[!t]
\begin{center}
\includegraphics[width=0.85\linewidth]{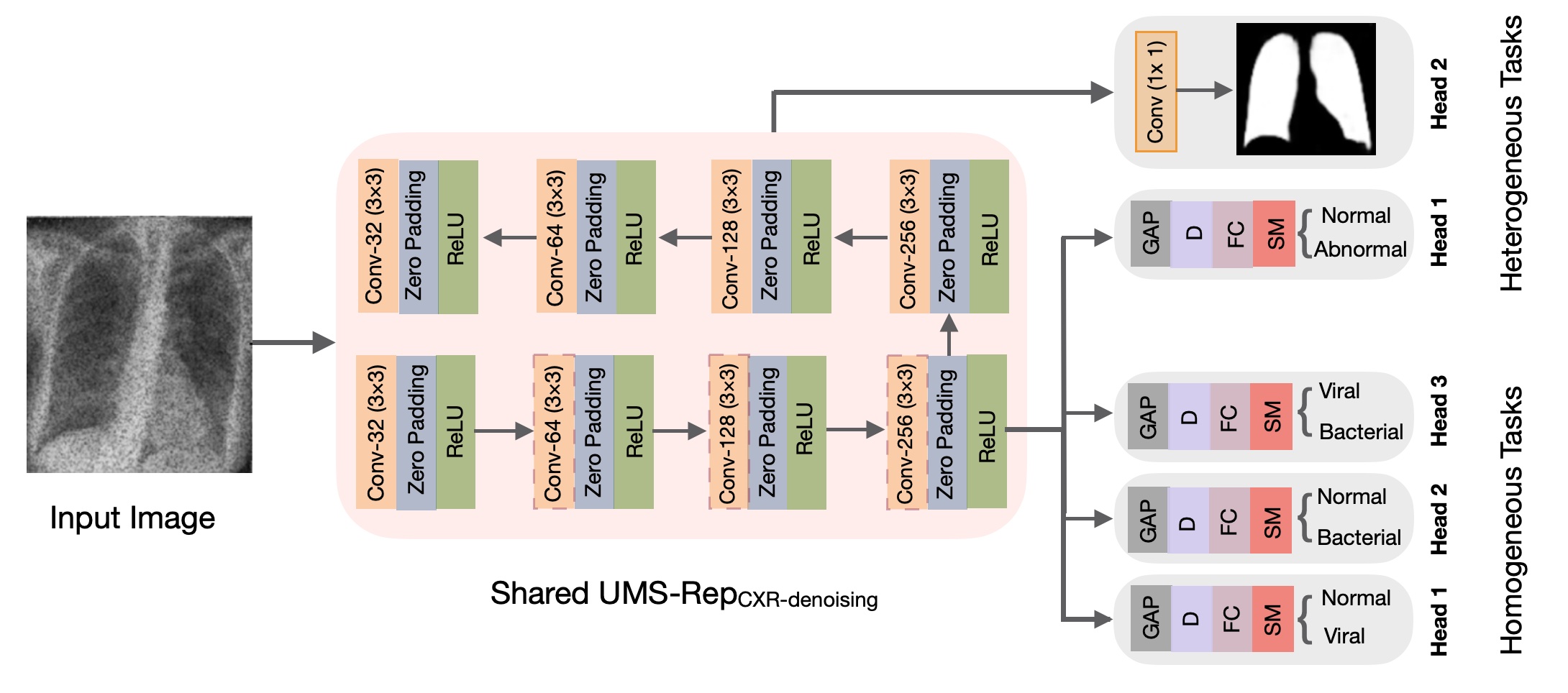}
\end{center}
\caption{UMS-Rep$_{CXR-Denoising}$ shared backbone and task-specific heads for heterogeneous and homogeneous tasks. GAP, D, FC, SM indicate global average pooling, dropout, fully connected, and softmax layers, respectively. Dashed orange boxes indicate convolutional layers with dilation. }
\label{fig:long}
\label{fig:onecol}
\end{figure*}


\begin{table*}[!t]
\centering
\begin{tabular}{|l|l|l|l|}
\hline
\textbf{Noise}  & \textbf{PSNR} & \textbf{SSIM} & \textbf{MS-SSIM} \\ \hline


Gaussian ($\sigma$=10) & 39.05dB & 0.96 & 0.99 \\ 
Gaussian ($\sigma$=20) & 35.38dB & 0.91 & 0.98 \\ 
Gaussian ($\sigma$=30) & 32.75dB & 0.89 & 0.95 \\ 
Gaussian ($\sigma$=40) & 30.54dB & 0.85 & 0.91 \\ 
Gaussian ($\sigma$=50) & 28.35dB & 0.79  & 0.88 \\ 
\hline
Poisson ($\mu$ = $\sigma^{2}$) & 33.37dB & 0.93 & 0.97 \\ 

\hline

\end{tabular}
\caption{Performance of customized UMS-Rep$_{CXR-Denoising}$ for image denoising. }
\end{table*}

\subsubsection{UMS-Rep Construction and Pre-processing}
UMS-Rep can be trained to learn CXR low-level features and pre-processing tasks using different learning paradigms. In this paper, we used unsupervised learning to train the UMS-Rep backbone for image denoising. Noise reduction is an essential pre-processing task that improves the quality of medical images while maintaining spatial resolution. The negative impact of image noise in subsequent tasks such as classification and segmentation has been reported in several works (e.g., \cite{cheng2010automated,mahmood2011comparison,nyma2012hybrid}). Further, the impact of different types of noise, such as Gaussian noise and speckle noise, on medical image segmentation has been reported in \cite{nyma2012hybrid}. Therefore, we propose to train a UMS-Rep that learns this important pre-processing task to enhance the performance of the subsequent target tasks.

A convolutional denoising autoencoder (CDAE or UMS-Rep$_{CXR-Denoising}$) was trained on the RSNA CXR dataset (see Table 1). The dataset was split at the patient-level into 70\%, 20\%, 10\% for training, validation, and testing, respectively. As depicted in Figure 6, UMS-Rep$_{CXR-Denoising}$ has four convolutional layers ($3 \times 3$) in the encoder to compress the input to its latent space representation. We used strided convolutions instead of max-pooling layers to increase the expressive capacity of the network, which would improve the overall performance without increasing the number of parameters as discussed in \cite{zeiler2010deconvolutional}. We used batch normalization layers to improve generalization and ReLU to speed up model training, resulting in faster convergence. The symmetrical decoder has upsampling layers to reconstruct the input from the latent space representation. We optimized CDAE (UMS-Rep$_{CXR-Denoising}$) to minimize the mean squared error (MSE) and reconstruct the input with minimal reconstruction error. The kernel size, stride, and optimizer for CDAE (UMS-Rep$_{CXR-Denoising}$) are 3, 2, and RMSprop, respectively. We trained the model with 16 batch size, an initial learning rate of $1 \times 10^{-3}$ that is reduced when the loss plateau, and early stopping. CDAE parameters were selected using Talos.

To generate noisy images for training, we added Gaussian noise with different ranges of standard deviations and Poisson noise to the images and used CDAE to reconstruct clean images. The reason behind adding the noise to the images is to demonstrate that UMS-Rep$_{CXR-Denoising}$ can learn to denoise Gaussian noise or any types of noise before sharing it among target tasks. Further, adding noise to the training process of UMS-Rep$_{CXR-Denoising}$ reduces overfitting and introduces regularization. Hence, we degraded the original images by Gaussian noise with different standard deviations ($\sigma$ = [10; 20; 30; 40; 50]) as well as Poisson noise ($\mu = \sigma^2$) and reported the performance in Table 2.

As shown in Table 2, we quantitatively measured the performance of reconstruction using the peak signal-to-noise ratio (PSNR), the structural similarity index (SSIM), and the multi-Scale structural similarity index (MS-SSIM). The typical range of PSNR is between 30 to 50 dB while the range of SSIM and MS-SSIM extends between -1 to +1, where higher is better (\cite{welstead1999fractal}). The PSNR, SSIM, and MS-SSIM values in Table 2 are computed for the test images. These results indicate that the trained UMS-Rep$_{CXR-Denoising}$ faithfully reconstructed the images and learned to extract useful CXR features while ignoring noise information. 

Once UMS-Rep$_{CXR-Denoising}$ is constructed, it is used as a shared representation among target CXR tasks with diverse or similar annotations as depicted in Figure 6. 


\subsubsection{Target Tasks Fine-tuning}
We used UMS-Rep$_{CXR-Denoising}$ as a shared encoder to simultaneously fine-tune target tasks with diverse and similar annotations. Examples of CXR target tasks with diverse annotations include lung segmentation (pixel-level) and abnormality classification (image-level). We used three tasks with image-level annotations as the homogeneous target tasks with similar annotations. These tasks are: (1) bacterial pneumonia vs normal classification, (2) viral pneumonia vs normal classification, and (3) bacterial vs viral pneumonia classification. 

Before fine-tuning the target tasks, task-specific layers (heads) are appended to UMS-Rep$_{CXR-Denoising}$ as presented in Figure 6. Note that in the case of the lung segmentation, we used the entire UMS-Rep$_{CXR-Denoising}$ and replaced the final convolutional layer with the one that has a single neuron to generate the binary lung masks. As for the classification tasks, only the encoder part of the optimized CDAE (UMS-Rep$_{CXR-Denoising}$) was instantiated and appended with the following layers: GAP, FC, D, and SM layers. 

After attaching task-specific heads to the shared UMS-Rep$_{CXR-Denoising}$, we experimented with three fine-tuning strategies to investigate their impact on the performance of the target tasks. In all experiments, the task-specific layers for the classification target tasks were fine-tuned to minimize the CCE loss while the task-specific layers for the segmentation tasks were fine-tuned to minimize the loss in equation 1. In the case of joint fine-tuning strategy, we summed the loss functions of the target tasks. The weights for task-specific losses are learned through empirical experiments (all weights are included in the code). In the case of alternating fine-tuning strategy, the loss for each target task is updated alternately as described in Section 2.3. For the independent fine-tuning strategy, we freeze the weights of UMS-Rep$_{CXR-Denoising}$ and independently minimize the loss function of the target tasks. 


\begin{table*}[!t]
\centering
\begin{tabular}{|l|l|l|l|l|l|l|l|l|l|}
\hline
\textbf{Method}  & \textbf{Tuning Strategy}  & \textbf{Target Task}  & \textbf{Accuracy} & \textbf{F-score} & \textbf{IoU} & \textbf{MCC}  \\ \hline

\hline

                 & Independent                                                     &   Lung Segmentation    &   \textbf{0.99} & \textbf{0.96} &  \textbf{0.98} & --
\\ 
 & Alternating &  Lung Segmentation    & 0.99 & 0.95 & 0.96 & -- \\

 & Joint &  Lung Segmentation    & 0.92 & 0.92 & 0.94 & --\\

 UMS-Rep$_{CXR-Denoising}$  &  &   & &  &  &  \\  

 & Independent &  Abnormality Classification    & \textbf{0.86} &  \textbf{0.83} & -- & \textbf{0.68} \\

 & Alternating &  Abnormality Classification   & 0.85 &  0.82 & -- & 0.66 \\

 & Joint &  Abnormality Classification    & 0.83 & 0.80 & -- & 0.63 \\ 
 
 &  &   & &  &  &  \\  
 
\hline
\hline
\multirow{2}{*}{Baseline} & Separate Model & Lung Segmentation & 0.97 & 0.91 & 0.95 & -- \\ 

 &  Separate Model & Abnormality Classification & 0.80 & 0.80 & -- & 0.60  \\ 
\hline

\end{tabular}
\caption{Performance of CXR target tasks with diverse annotations. Recall that we used three strategies to fine-tune the heads attached to UMS-Rep$_{CXR-Denoising}$. Bold values indicate highest performance.}
\end{table*}

\begin{table*}[!t]
\centering

\begin{tabular}{|l|l|l|l|l|}
\hline
\textbf{Method}               & \textbf{ Task} & \textbf{Computational Time} & \textbf{Training Parameters} 
\\ \hline
\multirow{3}{*}{Proposed} & Shared Source (UMS-Rep$_{CXR-Denoising}$)            & 302.20 minutes                 & 800,067          
\\  
                            & Lung Segmentation Head           & 8.11 minutes                & 786,497         
                            \\ 
                            & Abnormality Classification Head           & 2.51 minutes               & 295,715      
                            \\ 
\hline                        
  & Shared Source  \& 2 lightweight Heads        & 312.82 minutes               & 1,882,279       
  \\
  \hline
  \hline
\multirow{3}{*}{Baseline} & Separate Denoising Model &       302.20 minutes          & 800,067       
\\ 
 & Separate Lung Segmentation Model     &     148.02 minutes             & 786,497      
 \\
                            & Separate Abnormality Classification Model         &     9.25 minutes           & 1,573,506      
                            \\ 
                                        \hline

 &  3 Separate end-to-end Models for 3 Tasks        &     459.47 minutes           & 3,160,070      
 \\ 
\hline
\end{tabular}
\caption{Heterogeneous CXR target tasks: summary of computational time and training parameters for the proposed approach and the baseline approach. }
\end{table*}

\textit{Target CXR Tasks with Diverse Annotations:} 
These tasks were fine-tuned using the Shenzhen CXR dataset (see Table 1). We divided this dataset into different folds to perform 10-fold cross-validation. For testing, we used the Montgomery CXR dataset (see Table 1). The performance of fine-tuning lung segmentation and abnormality classification tasks based on the shared UMS-Rep$_{CXR-Denoising}$ is summarized in Table 3. As shown in the table, the independent fine-tuning strategy achieved slightly higher performance than the alternating strategy. However, the independent strategy increases the performance by a large margin as compared to the joint strategy. Statistically, the difference between the independent and alternating strategies is not significant using McNemar’s test (p $<$ 0.05). However, the difference between the independent and joint strategies is statistically significant (McNemar’s test, p $<$ 0.05). These results confirm the impact of the fine-tuning strategy on the performance and suggest that the joint strategy leads to lower performance when used with heterogeneous target tasks. This might be attributed to the negative transfer between irrelevant or conflicting tasks. Specifically, the classification task might transfer irrelevant information to the segmentation task or vice versa. Based on these results, we can conclude the superiority of independent strategy when fine-tuning heterogeneous tasks in CXR images.

We also compared the performance of the proposed approach with the baseline approach. To provide a fair comparison, CXR images are first denoised using a separate image denoising model (model 1), then we used the generated denoised images as input to the lung segmentation model (model 2) and abnormality classification model (model 3). These three models have the same architecture as in the proposed approach, but trained separately with random initialization. As shown in the last row of Table 3, the proposed approach outperformed the baseline approach in both target tasks. Statistically, the difference between the independent fine-tuning strategy and baseline approach is statistically significant (McNemar’s test, p $<$ 0.05). Note that the performance of the baseline segmentation model is higher than the joint fine-tuning strategy due to the negative transfer that might occur while minimizing the weighted sum of segmentation and classification tasks.

Table 4 shows the computational time and training parameters for the proposed and baseline approach. As shown in the table, the proposed approach significantly decreases the total training time and parameters as compared to the baseline approach, where separate models are used for separate tasks. Note that although the number of training parameters for the segmentation task does not change, the training time significantly decreased from 148.02 to 8.11 minutes. This demonstrates that the proposed approach, which uses the shared representation (UMS-Rep$_{CXR-Denoising}$) and its weight, leads to faster training convergence. Precisely, the weights of the shared UMS-Rep$_{CXR-Denoising}$ resulted in improved initialization and faster learning/convergence for the target tasks.




\begin{table*}[!t]
\centering
\begin{tabular}{|l|l|l|l|l|l|}
\hline
\textbf{Method}  & \textbf{Tuning Strategy}  & \textbf{Target Task}  & \textbf{Accuracy} & \textbf{F-score}  & \textbf{MCC}  \\ \hline

\hline

                 & Independent                                                    &    Classification (Bacterial vs Normal)
    &   0.95 & 0.95 &  0.90 
\\  
 & Alternating &  Classification (Bacterial vs Normal)    & \textbf{0.96} & \textbf{0.96} & \textbf{0.92}
\\
 & Joint &  Classification (Bacterial vs Normal)    & 0.96 & 0.95 & 0.91\\ 

& & & & &  \\
 & Independent &  Classification (Viral vs Normal)    & 0.95 &  0.92 & 0.88 \\  

{UMS-Rep$_{CXR-Denoising}$} & Alternating &  Classification (Viral vs Normal)   & 0.96 &  0.93 & 0.90  \\

 & Joint  &  Classification (Viral vs Normal)   & 0.\textbf{97} & \textbf{0.94} & \textbf{0.92}   \\ 

& & & & &  \\
 & Independent &  Classification (Bacterial vs Viral)    & 0.81 &  0.76 & 0.56 \\  
 
 & Alternating &  Classification (Bacterial vs Viral)   & 0.82 &  0.77 & 0.58  \\ 
 
  & Joint &  Classification (Bacterial vs Viral)   & \textbf{0.83} & \textbf{0.80} & \textbf{0.62} \\ 
 & & & & & \\
\hline
\hline
\multirow{3}{*}{Baseline} & Separate Model & Classification (Bacterial vs Normal) & 0.94 & 0.94 & 0.89 \\

 & Separate Model & Classification (Viral vs Normal) & 0.93 & 0.90 & 0.86 \\

 & Separate Model & Classification (Bacterial vs Viral)  & 0.80 & 0.75 & 0.54  \\ 
 
\hline

\end{tabular}
\caption{Performance of homogeneous CXR target tasks using the proposed method and the baseline. Recall that we used three strategies to fine-tune the heads attached to UMS-Rep$_{CXR-Denoising}$. Bold values indicate best performance. }
\end{table*}

\begin{table*}[!t]
\centering

\begin{tabular}{|l|l|l|l|l|}
\hline
\textbf{Method}               & \textbf{ Task} & \textbf{Computational Time} & \textbf{Training Parameters} 
\\ \hline
 \multirow{3}{*}{Proposed}                           &  Classification Head 1 (Bacterial vs Normal)           & 1.90 minutes                & 295,715        
                            \\ 
                            &  Classification Head 2 (Viral vs Normal)          & 1.75 minutes                & 295,715         
                            \\ 
                                & Classification Head 3 (Bacterial vs Viral)       & 1.16 minutes                & 295,715       
                                \\ 
\hline         
      & Shared Source \& 3 lightweight Heads       & 4.81 minutes                & 887,145        
      \\ 
\hline 
\hline 

\multirow{2}{*}{Baseline}                            & Separate Model (Bacterial vs Normal)           &     12.75 minutes           & 1,573,506       
                            \\ 
                                        & Separate Model (Viral vs Normal)       &     9.21 minutes        & 1,573,506       
                                        \\
                                          & Separate Model (Bacterial vs Viral)       &     12.19 minutes           & 1,573,506         
                                          \\ 
                                          \hline             &  Separate end-to-end Models for 4 Tasks         &     34.15 minutes           & 4,720,518     
                                          \\ 
                                          \hline

\end{tabular}
\caption{Homogeneous CXR target tasks: summary of computational time and training parameters for the proposed and baseline approaches. }
\end{table*}

\textit{Target CXR Tasks with Similar Annotations:} 
These tasks were fine-tuned using the train set of the Pediatric pneumonia dataset (Table 1), which was further divided into multiple folds to perform 10-fold cross-validation analysis. We then used the hold-out test set for reporting the performance. Recall that this dataset contains three classes: normal, bacterial pneumonia, and viral pneumonia. Examples of these classes are given in Figure 2.

Table 5 shows the performance of fine-tuning the classification tasks based on the shared UMS-Rep$_{CXR-Denoising}$. As shown in the table, the joint fine-tuning strategy achieved higher performance than both the alternating and independent strategies in most cases. Statistically, the difference in performance between the joint and independent strategies is significant (McNemar’s test, p $<$ 0.05), except for bacterial vs normal. These results suggest the superiority of the joint strategy for fine-tuning medical image tasks with diverse annotations. This can be explained by the ability of the joint loss function to capture the task differences and implicitly model the task relationships; i.e., the inter-task relationships among target task 1 (bacterial vs normal), target task 2 (viral vs normal), and target task 3 (bacterial vs viral) (all use image-label) are better captured using the joint strategy. Similarly, the alternating fine-tuning strategy achieved better performance than the independent strategy as it allows, by alternately optimizing $\Theta_{src}$ and $\Theta_{T_i}$, to transfer some of the information from each task to the other. These results suggest the superiority of joint and alternating strategies for cooperative medical image tasks with similar annotations. 

We also compared the performance of the proposed approach with the baseline, where separate end-to-end models with random initialization are used for separate target tasks. Specifically, we first used a separate image denoising model to provide a fair comparison with our approach. We then used the output denoised images as input to three separate models, one for each target task. These three models have the same architectures as in the proposed approach, but trained separately with random initialization. As shown in the last row of Table 5, the proposed approach outperformed the baseline approach in all target tasks. Statistically, the difference in performance between the joint fine-tuning strategy and baseline approach is significant for all tasks (p $<$ 0.05). These results indicate that the proposed approach, by transferring the knowledge from a shared modality-specific source to target tasks, can improve the generalizability and lead to better performance as compared to the baseline approach. Further, the results suggest that the joint fine-tuning of similar tasks can enhance the overall performance while reducing the learning time as shown in Table 6. 

Table 6 shows the computational time and training parameters for the proposed and baseline approach. As shown in the table, the proposed approach significantly decreases the total training time and parameters as compared to the baseline approach, where separate models (encoder-decoder) are used for separate tasks. Specifically, the proposed approach reduces the number of training parameters by $\sim$ 81\% as compared to the baseline approach. Similarly, the proposed approach leads to a reduction by 86\% in the computational time, and faster convergence. Finally, it is important to note that no matter the number of target tasks, our approach would lead to lower computational times and parameters as compared to the baseline approach. This is attributed to the fact that it 1) uses a single shared encoder and $N$ lightweight task-specific decoders instead of $N$ encoder-decoder for $N$ tasks and 2) the weights of the shared UMS-Rep improved initialization and leads to faster convergence for all target tasks. 

\begin{figure}[!b]
\begin{center}

\includegraphics[width=0.45\linewidth]{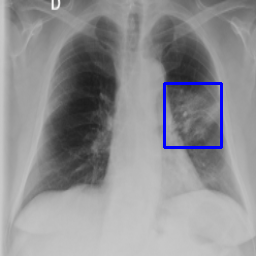}
\includegraphics[width=0.45\linewidth]{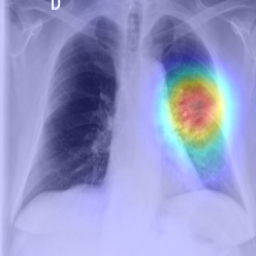}

\end{center}
\caption{Visual explanation task. Left: original image with GT annotation (bounding box in blue). Right: visualization on a testing image. High activation (red pixels) is observed in the affected ROIs. }
\label{fig:long}
\label{fig:onecol}
\end{figure}

\subsubsection{Derivable Task} 
After fine-tuning the target classification task, we can visualize the output of the target task by simply computing the gradient of the winning class with respect to the last convolutional layer of the target task as described in \cite{chen2018gradnorm}. Then, we compute the average, weigh it against the output of this layer, and normalize it between 0 and 1 to generate the heatmap. This map visualizes discriminative areas the head looks at when classifying into normal and abnormal. As this visualization task relies on the output of the classification head, we can define it as a task derivable from the target task. Figure 7 presents the visualization of the abnormality classification task.

\subsection{Doppler echo Imaging Modality}
Although the high-level (i.e., task-specific) features of Doppler echo target tasks might differ, they have relatively similar low-level features. Hence, a single UMS-Rep can be trained to learn these low-level features and then shared among  different target tasks that learn task-specific features.

\subsubsection{UMS-Rep Construction}
In this evaluation, we used supervised learning to construct the UMS-Rep on Doppler echo dataset using the annotations of the flow classification task (see Table 1). We used 70\% of the dataset to construct the UMS-Rep. As shown in Figure 8, this UMS-Rep has six convolutional layers ($3 \times 3$) with the same padding. Dilated kernels (size 2) were used in the fourth, fifth, and sixth convolutional layers to capture wider context at a reduced computational cost. We used ReLU after each convolutional layer to speed-up model training and convergence. The output of the deepest convolutional layer from the optimized CNN was fed to the GAP and FC layers. To reduce overfitting, the output of the FC layer was fed to a dropout layer (0.5). The last FC layer has three neurons corresponding to three flow classes: TR, MV, and MA. We used Talos to select the optimal parameters from the following ranges: kernel size [3, 5, 7], dilation rate [2, 3], dropout ratio [0.1, 0.3, 0.5], optimizer [SGD, Adam, RMSprop], and batch size [16, 32, 64]. Talos outputs 3, 2, 0.5, Adam, and 16 for kernel size, dilation rate, dropout ratio, optimizer, and batch size, respectively. We used 64 epochs and optimized the model to minimize the CCE loss.

Once the optimized UMS-Rep$_{echo}$ is constructed, it is used as a shared encoder among target echo tasks with diverse and similar annotations as depicted in Figure 8. 


\begin{figure*}[!t]
\begin{center}
\includegraphics[width=0.90\linewidth]{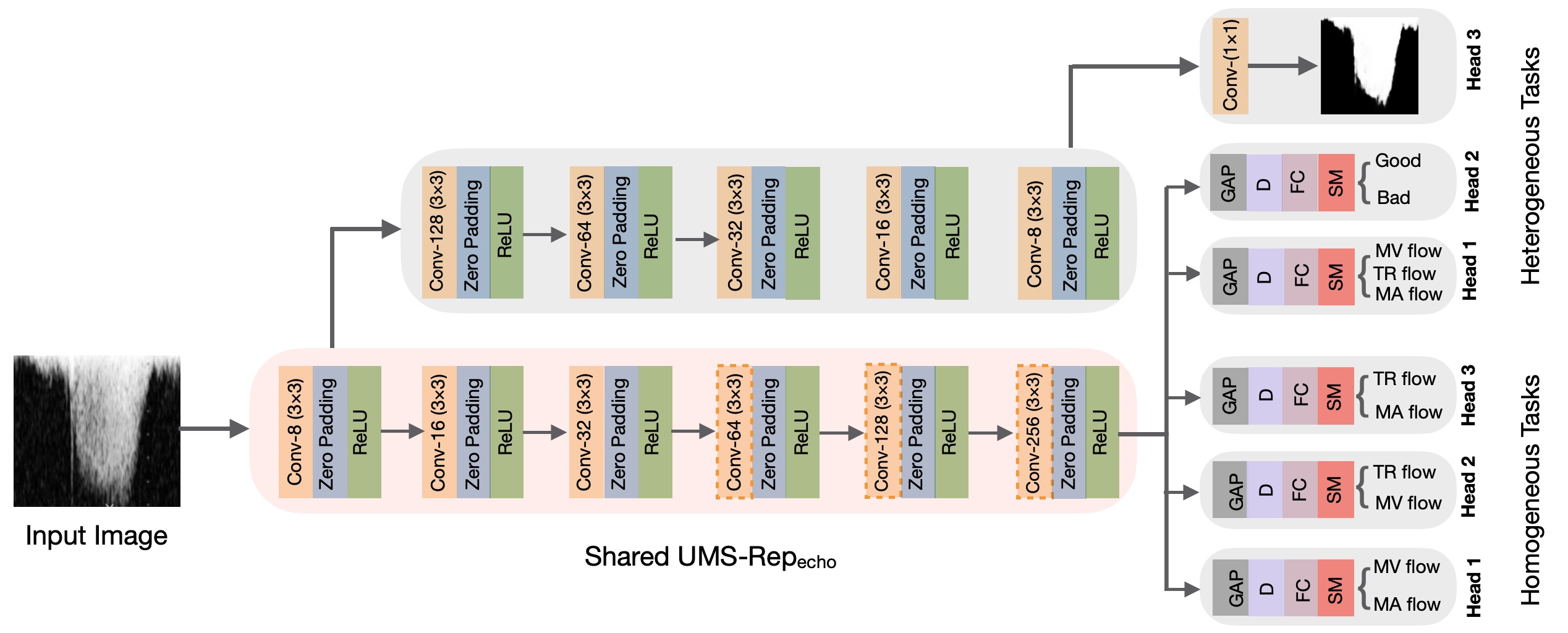}

\end{center}
\caption{UMS-Rep$_{echo}$ (pink box) is shared among target tasks. GAP, D, FC, SM indicate global average pooling, droput, fully connected, and softmax layers, respectively. Dashed orange boxes indicate convolutional layers with dilation.}

\label{fig:long}
\label{fig:onecol}
\end{figure*}

\subsubsection{Target Tasks Fine-tuning}
We used UMS-Rep$_{echo}$ as a shared backbone to simultaneously fine-tune target tasks with diverse and similar annotations. Examples of echo Doppler tasks with diverse annotations include envelope segmentation (pixel-level), flow classification (image-level), and image quality assessment (image-level). 
We used three classification tasks with image-level annotations as the homogeneous target tasks with similar annotations. These classification tasks are: (1) TR vs MV, (2) TR vs MA, and (3) MV vs MA.

We fine-tuned the target tasks using the remaining 30\% of echo Doppler dataset, where 20\% was used for training and validation (10-folds cross-validation) and 10\% was used for testing. Before fine-tuning the target tasks, task-specific layers (heads) are appended to UMS-Rep$_{echo}$ as presented in Figure 8. For the envelope segmentation task, the UMS-Rep$_{echo}$ was truncated at the deepest convolutional layer, and a symmetrical decoder was appended as shown in Figure 8. This head was fine-tuned using Adam optimizer to optimize the loss in equation 1. To create the task speciﬁc-layers (heads) for the classification tasks, we appended GAP, FC, D, and SM layers to UMS-Rep$_{echo}$ as shown in Figure 8. The heads for the classification tasks were fine-tuned using SGD optimizer to minimize the CCE loss. After attaching task-specific heads to the shared UMS-Rep$_{echo}$, we experimented with three fine-tuning strategies to investigate their impact on the performance of the target tasks.

\begin{table*}[!t]
\centering
\begin{tabular}{|l|l|l|l|l|l|l|l|l|l|}
\hline
\textbf{Method}  & \textbf{Tuning Strategy}  & \textbf{Target Task}  & \textbf{Accuracy} & \textbf{F-score} & \textbf{IoU} & \textbf{MCC}  \\ \hline

\hline

                 & Independent                                                      &   Envelope Segmentation    &   \textbf{0.97} & \textbf{0.95} &  \textbf{0.98} & --
\\ 
 & Alternating &  Envelope Segmentation    & 0.96 & 0.94 & 0.98 & -- \\

 & Joint &  Envelope Segmentation    & 0.91 & 0.93 & 0.94 & --\\

& & & & & & \\
 & Independent &  Quality Assessment    & \textbf{0.97} &  \textbf{0.96} & -- & \textbf{0.92} \\

UMS-Rep$_{echo}$  & Alternating &  Quality Assessment    & 0.95 &  0.91 & -- & 0.87 \\

 & Joint  &  Quality Assessment     & 0.93 & 0.88 & -- & 0.83 \\ 
 
 &  &   & &  &  &  \\  
 
  & Independent &  Flow Classification     & \textbf{0.93} & \textbf{0.91} & -- & \textbf{0.85} \\ 

 & Alternating &  Flow Classification     & 0.91 & 0.90 & -- & 0.82 \\ 
 
  & Joint  &  Flow Classification     & 0.92 & 0.90 & -- & 0.84 \\ 
  
 &  &   & &  &  &  \\
 
\hline
\hline
\multirow{2}{*}{Baseline} & Separate Model & Envelope Segmentation & 0.93 & 0.92 & 0.95 & -- \\ 

 &  Separate Model & Quality Assessment  & 0.88 & 0.85 & -- & 0.78  \\ 
 &  Separate Model & Flow Classification  & 0.91 & 0.88 & -- & 0.81  \\ 
\hline

\end{tabular}
\caption{Performance of heterogeneous echo target tasks using the proposed method and the baseline. Bold values indicate best performance. }
\end{table*}

\begin{table*}[!t]
\centering

\begin{tabular}{|l|l|l|l|l|}
\hline
\textbf{Method}               & \textbf{ Task} & \textbf{Computational Time} & \textbf{Training Parameters} 
\\ \hline
\multirow{4}{*}{Proposed}                             &  Shared UMS-Rep$_{echo}$            & 28.19 minutes                & 460,115        
\\ 
                          & Envelope Segmentation Head           & 5.80 minutes                & 786,497        
\\ 
                            & Quality Assessment Head         & 3.00 minutes               & 295,715       
                            \\ 
                              & Flow Classification Head           & 1.78 minutes               & 295,715       
                              \\ 
\hline                        
  & Shared UMS-Rep$_{echo}$  \& 3 lightweight Heads        & 38.77 minutes               & 1,838,042      
  \\
  \hline
  \hline
\multirow{3}{*}{Baseline} & Separate Envelope Segmentation Model &       130.02 minutes          & 786,497     
\\ 
 & Separate Quality Assessment     &     29.34 minutes             & 1,573,506     
 \\
                            & Separate Flow Classification Model         &     36.27 minutes           & 1,573,506     
                            \\ 
                                        \hline

 &  3 Separate end-to-end Models for 3 Tasks        &     195.63 minutes           & 3,933,509      
 \\ 
\hline
\end{tabular}
\caption{Heterogeneous echo target tasks: summary of time and training parameters for the proposed approach and baseline. }
\end{table*}

\textit{Target Echo Tasks with Diverse Annotations:} 
Table 7 shows the performance of fine-tuning envelope segmentation, flow classification, and quality assessment tasks. Observe that the independent strategy increases the performance by a large margin as compared to the joint strategy. Statistically, the difference between the independent and joint strategies is significant (McNemar’s test, p $<$ 0.05). These results suggest the superiority of independent strategy when fine-tuning echo tasks with diverse annotations. We also compared the performance of the proposed approach with the baseline approach. In the baseline approach, the segmentation model as well as the quality assessment and flow classification models have the same architecture as in the proposed approach, but trained separately with random initialization. As shown in the last row of Table 7, the proposed approach outperformed the baseline approach. The difference between the independent fine-tuning strategy and baseline approach is statistically significant (McNemar’s test, p $<$ 0.05). Again, the joint fine-tuning strategy does not improve the performance of the segmentation task. We believe this could be interpreted as being a result of the negative transfer among irrelevant tasks while learning a single joint function.

Table 8 shows the computational time and training parameters for the proposed and baseline approach. As shown in the table, the proposed approach significantly decreases the total computational time ($\downarrow \sim$ 80\%) and training parameters ($\downarrow \sim$  53\%) as compared to the baseline approach, where separate models are used for separate tasks. Interestingly, although the number of training parameters for the segmentation task does not change, the training time decreased from 130.02 to 5.80 minutes ($\sim$ 96\%) as a result of using UMS-Rep$_{echo}$ backbone. Precisely, the weights of the shared UMS-Rep$_{echo}$ resulted in improved initialization and faster convergence. These results demonstrate the benefits of using UMS-Rep$_{echo}$ as a shared backbone.

\begin{table*}[!t]
\centering
\begin{tabular}{|l|l|l|l|l|l|}
\hline
\textbf{Method}  & \textbf{Tuning Strategy}  & \textbf{Target Task}  & \textbf{Accuracy} & \textbf{F-score}  & \textbf{MCC}  \\ \hline

\hline

                 & Independent                                                      &    Classification (MV vs MA)
    &  0.92 & 0.91 &  0.83 
\\
 & Alternating &  Classification (MV vs MA)    & 0.94 & 0.94 & 0.87
\\

 & Joint &  Classification (MV vs MA)    & \textbf{0.95} & \textbf{0.95} & \textbf{0.89} \\ 

& & & & & \\
 & Independent  &  Classification (MV vs TR)    & 0.95 &  0.94 & 0.88 \\  

UMS-Rep$_{echo}$  & Alternating  &  Classification (MV vs TR)   & 0.96 &  0.96 & 0.91  \\
 

 & Joint  &  Classification (MV vs TR)   & \textbf{0.97} & \textbf{0.96} & \textbf{0.93}   \\ 

& & & & & \\

 & Independent &  Classification (MA vs TR)    & 0.97 &  0.96 & 0.92 \\  
 
  & Alternating &  Classification (MA vs TR)   & 0.98 &  0.96 & 0.93  \\ 
 
  & Joint &  Classification (MA vs TR)   & \textbf{0.98} & \textbf{0.98} & \textbf{0.95} \\ 
 & & & & & \\

\hline
\hline
\multirow{3}{*}{Baseline} & Separate Model & Classification (MV vs MA) & 0.91 & 0.91 & 0.81 \\

 & Separate Model & Classification (MV vs TR) & 0.94 & 0.93 & 0.86 \\ 
 

 & Separate Model & Classification (MA vs TR)  & 0.98 & 0.96 & 0.93  \\ 
\hline

\end{tabular}
\caption{Performance of homogeneous echo target tasks using the proposed method and the baseline. Bold values indicate the best performance.}
\end{table*}

\begin{table*}[!h]
\centering

\begin{tabular}{|l|l|l|l|l|}
\hline
\textbf{Method}               & \textbf{ Task} & \textbf{Computational Time} & \textbf{Training Parameters} 
\\ \hline
\multirow{4}{*}{Proposed} & Shared Source (UMS-Rep$_{echo}$)            & 28.19 minutes                 & 460,115                      \\  
                             &  Classification Head 1 (MV vs MA)           & 2.42 minutes                & 295,715        
\\ 
                            &  Classification Head 2 (MV vs TR)          & 1.25 minutes                & 295,715        
                            \\ 
                                & Classification Head 3 (MA vs TR)       & 1.83 minutes                & 295,715       
                                \\ 
\hline         
      & 1 Shared Source \& 3 lightweight Heads       & 33.69 minutes                & 1,347,260        
      \\ 
\hline 
\hline 

\multirow{2}{*}{Baseline} & Separate Model (MV vs MA)           &     16.67 minutes           & 1,573,506      
\\ 
                            & Separate Model (MV vs TR)           &     10.85 minutes           & 1,573,506       
                            \\ 
                                        & Separate Model (MA vs TR)       &     9.25 minutes        & 1,573,506      
                                        \\
                         
                                          \hline             & 3 Separate end-to-end Models for 3 Tasks         &     36.77 minutes           & 4,720,518     
                                          \\ 
                                          \hline

\end{tabular}
\caption{Homogeneous echo target tasks: computational time and training parameters for the proposed and baseline approaches. }
\end{table*}

\textit{Target Echo Tasks with Similar Annotations:} 
Table 9 shows the performance of fine-tuning the classification tasks based on the shared UMS-Rep$_{echo}$. We can conclude from the table that the joint fine-tuning strategy achieved higher performance in most cases. Although the difference in the performance between the joint and alternating strategies is not significant in most cases, it is statistically significant between the joint and independent (McNemar’s test, p $<$ 0.05). These results suggest that optimizing a single joint loss function for homogeneous target tasks leads to better overall performance. In other words, the results suggest the superiority of the joint strategy for fine-tuning medical image tasks with similar annotations as it has the ability to capture the similarities and differences among similar tasks. We also compared the performance of the proposed approach with the baseline approach. As shown in the last row of Table 9, the proposed approach outperformed the baseline approach. The statistical difference between the joint fine-tuning strategy and baseline approach is significant (p $<$ 0.05). Also, the proposed approach decreases the total training parameters and computational time, as compared to the baseline approach (Table 10). Although our approach decreases the computational time in Table 10 only slightly ($\sim$ 8\%), it is important to note that the reduction in the computational time will continue as we add more target tasks; e.g., adding another classification task would make the total computational time for our approach and baseline 34.94 (33.69 + 1.25) minutes and 46.02 (36.77 + 9.25) minutes respectively, and lead to $\sim$ 24\% reduction. This indicates that our approach will always have lower computations as it involves sharing a single decoder among target tasks, which enhances initialization and leads to faster convergence. 


\subsubsection{Derivable Task}
After learning the target tasks, we derived a recommendation task. This derivable task is generated based on combining the outputs of the flow classification and quality assessment tasks. This recommendation is used to decide if the image is suitable for further analysis. Currently, the echocardiographer manually excludes low-quality TR flow images with unclear envelopes from further analysis because they decrease the accuracy of measurements. Figure 9 shows examples of the recommendation task. 

\begin{figure}[!t]
\begin{center}

\includegraphics[width=0.45\linewidth]{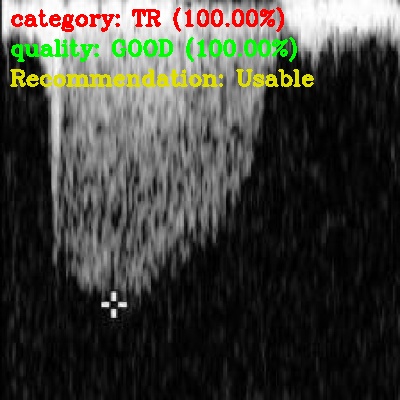}
\includegraphics[width=0.45\linewidth]{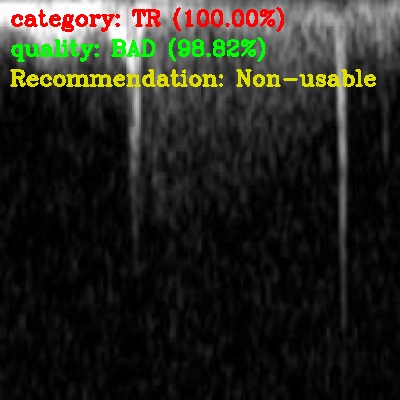}

\caption{Derivable recommendation task. Examples of merging the outputs of tasks to select only good quality TR images. }

\end{center}

\label{fig:long}
\label{fig:onecol}
\end{figure}


\subsection{Discussion}
To resolve the challenges posed by applying the traditional training approach on medical images, we proposed a novel sequence of training that involves constructing UMS-Rep followed by using it as a shared backbone to simultaneously fine-tune target tasks with similar or diverse annotations. 

Our experimental results show that the proposed approach improved the generalization, and performance (up to 9\%) of target tasks by transferring the knowledge from a modality-specific shared source to these tasks. Further, using a single representation to fine-tune multiple target tasks reduced the computations by a large margin and prevented unnecessary repetitions of training individual models for individual pre-processing and visual analysis tasks; i.e., single encoder and $N$ lightweight task-specific heads instead of $N$ encoder-decoder models for $N$ tasks. Our results suggest the superiority of the independent fine-tuning strategy for heterogeneous medical image tasks, and the joint strategy (followed by alternating) for homogeneous medical image tasks. This can be explained by the ability of the joint loss function to capture the task differences and implicitly model inter-task relationships between similar tasks. Similarly, the alternating strategy allows discovering some of commonality between target tasks. As these results are consistent in two imaging modalities (CXR and echo Doppler), we may conclude that independent strategy should be used to fine-tune heterogeneous target tasks and joint (or alternating) strategy should be used to fine-tune homogeneous target tasks. Finally, it is important to note the benefit of the proposed sequence of training in reducing overfitting as reported in (\cite{baxter1997bayesian}). 

The proposed approach can improve the efficiency and performance of medical image tasks as follows. First, transferring the knowledge from shared modality representation (which can be trained without labels) to the segmentation task improves generalization and leads to faster convergence for the target tasks. Second, our results demonstrated that the fine-tuning strategy greatly impacts the performance of the target task, and suggested to avoid using joint strategy for fine-tuning conflicting tasks. In other words, independent fine-tuning using a shared modality-specific representation leads to better performance when used with heterogeneous tasks while joint or alternating fine-tuning leads to better performance when used with homogeneous or similar tasks.

Although we only demonstrated the feasibility of our approach to learn a single pre-processing task (i.e., noise reduction) and some relevant tasks in medical image analysis, the proposed approach is flexible and can be easily extended to integrate other pre-processing tasks (e.g., super resolution \cite{zamzmi2020accelerating}) and learn any number of target tasks. For example, a bounding box regression task can be added to the UMS-Rep by appending a task-specific head that has region pooling layers for extracting region-wise features and FC layers for box classification and regression. The proposed approach can also be extended to analyze 3D images using a 3D CNN as the shared representation and task-specific layers. 
 
\section{Conclusion}
In this paper, we proposed a unified modality-specific approach under the MTL framework where the encoder is shared across different target tasks with diverse and similar annotations. We applied the proposed approach to two medical imaging modalities, namely CXR and echo Doppler. We also explored different strategies for fine-tuning the target tasks to investigate the impact of the utilized strategy on their performance. To the best of our knowledge, the problem of learning to simultaneously transfer knowledge from shared representation to multiple target tasks has seldom been studied in medical images. Our experiments show that the proposed approach can improve the generalization and performance of target tasks, while providing computational efficiency.

\section*{Competing interests}
The authors declare no competing interests.

\section*{Acknowledgments}
This research is supported by the Intramural Research Program of the National Library of Medicine (NLM) and the National Institutes of Health (NIH).




\printcredits

\bibliographystyle{cas-model2-names}

\bibliography{cas-refs}

\begin{thebibliography}{31}
\expandafter\ifx\csname natexlab\endcsname\relax\def\natexlab#1{#1}\fi
\providecommand{\url}[1]{\texttt{#1}}
\providecommand{\href}[2]{#2}
\providecommand{\path}[1]{#1}
\providecommand{\DOIprefix}{doi:}
\providecommand{\ArXivprefix}{arXiv:}
\providecommand{\URLprefix}{URL: }
\providecommand{\Pubmedprefix}{pmid:}
\providecommand{\doi}[1]{\href{http://dx.doi.org/#1}{\path{#1}}}
\providecommand{\Pubmed}[1]{\href{pmid:#1}{\path{#1}}}
\providecommand{\bibinfo}[2]{#2}
\ifx\xfnm\relax \def\xfnm[#1]{\unskip,\space#1}\fi
\bibitem[{Bai et~al.(2019)}]{bai2019self}
\bibinfo{author}{Bai, W.}, et~al., \bibinfo{year}{2019}.
\newblock \bibinfo{title}{Self-supervised learning for cardiac mr image
  segmentation by anatomical position prediction}, in:
  \bibinfo{booktitle}{International Conference on Medical Image Computing and
  Computer-Assisted Intervention}, \bibinfo{organization}{Springer}. pp.
  \bibinfo{pages}{541--549}.
\bibitem[{Balakrishnan et~al.(2018)}]{balakrishnan2018unsupervised}
\bibinfo{author}{Balakrishnan, G.}, et~al., \bibinfo{year}{2018}.
\newblock \bibinfo{title}{An unsupervised learning model for deformable medical
  image registration}, in: \bibinfo{booktitle}{Proceedings of the IEEE
  conference on computer vision and pattern recognition}, pp.
  \bibinfo{pages}{9252--9260}.
\bibitem[{Baxter(1997)}]{baxter1997bayesian}
\bibinfo{author}{Baxter, J.}, \bibinfo{year}{1997}.
\newblock \bibinfo{title}{A bayesian/information theoretic model of learning to
  learn via multiple task sampling}.
\newblock \bibinfo{journal}{Machine learning} \bibinfo{volume}{28},
  \bibinfo{pages}{7--39}.
\bibitem[{Chen et~al.(2018a)}]{chen2018multi}
\bibinfo{author}{Chen, C.}, et~al., \bibinfo{year}{2018}a.
\newblock \bibinfo{title}{Multi-task learning for left atrial segmentation on
  ge-mri}, in: \bibinfo{booktitle}{International workshop on statistical
  atlases and computational models of the heart},
  \bibinfo{organization}{Springer}. pp. \bibinfo{pages}{292--301}.
\bibitem[{Chen et~al.(2018b)}]{chen2018gradnorm}
\bibinfo{author}{Chen, Z.}, et~al., \bibinfo{year}{2018}b.
\newblock \bibinfo{title}{Gradnorm: Gradient normalization for adaptive loss
  balancing in deep multitask networks}, in: \bibinfo{booktitle}{International
  Conference on Machine Learning}, pp. \bibinfo{pages}{793--802}.
\bibitem[{Cheng et~al.(2010)}]{cheng2010automated}
\bibinfo{author}{Cheng, H.D.}, et~al., \bibinfo{year}{2010}.
\newblock \bibinfo{title}{Automated breast cancer detection and classification
  using ultrasound images: A survey}.
\newblock \bibinfo{journal}{Pattern recognition} \bibinfo{volume}{43},
  \bibinfo{pages}{299--317}.
\bibitem[{Christodoulidis et~al.(2016)}]{christodoulidis2016multi}
\bibinfo{author}{Christodoulidis, A.}, et~al., \bibinfo{year}{2016}.
\newblock \bibinfo{title}{A multi-scale tensor voting approach for small
  retinal vessel segmentation in high resolution fundus images}.
\newblock \bibinfo{journal}{Comput. Med. Imaging Graph.} \bibinfo{volume}{52},
  \bibinfo{pages}{28--43}.
\bibitem[{Dong et~al.(2015)}]{dong2015multi}
\bibinfo{author}{Dong, D.}, et~al., \bibinfo{year}{2015}.
\newblock \bibinfo{title}{Multi-task learning for multiple language
  translation}, in: \bibinfo{booktitle}{Proceedings of the 53rd Annual Meeting
  of the Association for Computational Linguistics and the 7th International
  Joint Conference on Natural Language Processing (Volume 1: Long Papers)}, pp.
  \bibinfo{pages}{1723--1732}.
\bibitem[{Elhoseiny et~al.(2015)}]{elhoseiny2015convolutional}
\bibinfo{author}{Elhoseiny, M.}, et~al., \bibinfo{year}{2015}.
\newblock \bibinfo{title}{Convolutional models for joint object categorization
  and pose estimation}.
\newblock \bibinfo{journal}{arXiv preprint arXiv:1511.05175} .
\bibitem[{Gulati et~al.(2020)}]{gulati2020application}
\bibinfo{author}{Gulati, T.}, et~al., \bibinfo{year}{2020}.
\newblock \bibinfo{title}{Application of an enhanced deep super-resolution
  network in retinal image analysis}, in: \bibinfo{booktitle}{Ophthalmic
  Technologies XXX}, \bibinfo{organization}{International Society for Optics
  and Photonics}. p. \bibinfo{pages}{112181K}.
\bibitem[{Jaeger et~al.(2014)}]{jaeger2014two}
\bibinfo{author}{Jaeger}, et~al., \bibinfo{year}{2014}.
\newblock \bibinfo{title}{Two public chest x-ray datasets for computer-aided
  screening of pulmonary diseases}.
\newblock \bibinfo{journal}{Quantitative imaging in medicine and surgery}
  \bibinfo{volume}{4}, \bibinfo{pages}{475}.
\bibitem[{Kendall et~al.(2018)}]{kendall2018multi}
\bibinfo{author}{Kendall, A.}, et~al., \bibinfo{year}{2018}.
\newblock \bibinfo{title}{Multi-task learning using uncertainty to weigh losses
  for scene geometry and semantics}, in: \bibinfo{booktitle}{Proceedings of the
  IEEE Conference on Computer Vision and Pattern Recognition}, pp.
  \bibinfo{pages}{7482--7491}.
\bibitem[{Kermany et~al.(2018)Kermany, Goldbaum, Cai, Valentim, Liang, Baxter,
  McKeown, Yang, Wu, Yan et~al.}]{kermany2018identifying}
\bibinfo{author}{Kermany, D.S.}, \bibinfo{author}{Goldbaum, M.},
  \bibinfo{author}{Cai, W.}, \bibinfo{author}{Valentim, C.C.},
  \bibinfo{author}{Liang, H.}, \bibinfo{author}{Baxter, S.L.},
  \bibinfo{author}{McKeown, A.}, \bibinfo{author}{Yang, G.},
  \bibinfo{author}{Wu, X.}, \bibinfo{author}{Yan, F.}, et~al.,
  \bibinfo{year}{2018}.
\newblock \bibinfo{title}{Identifying medical diagnoses and treatable diseases
  by image-based deep learning}.
\newblock \bibinfo{journal}{Cell} \bibinfo{volume}{172},
  \bibinfo{pages}{1122--1131}.
\bibitem[{Kisilev et~al.(2016)Kisilev, Sason, Barkan and
  Hashoul}]{kisilev2016medical}
\bibinfo{author}{Kisilev, P.}, \bibinfo{author}{Sason, E.},
  \bibinfo{author}{Barkan, E.}, \bibinfo{author}{Hashoul, S.},
  \bibinfo{year}{2016}.
\newblock \bibinfo{title}{Medical image description using multi-task-loss cnn},
  in: \bibinfo{booktitle}{Deep Learning and Data Labeling for Medical
  Applications}. \bibinfo{publisher}{Springer}, pp. \bibinfo{pages}{121--129}.
\bibitem[{Litjens et~al.(2017)}]{litjens2017survey}
\bibinfo{author}{Litjens, G.}, et~al., \bibinfo{year}{2017}.
\newblock \bibinfo{title}{A survey on deep learning in medical image analysis}.
\newblock \bibinfo{journal}{Medical image analysis} \bibinfo{volume}{42},
  \bibinfo{pages}{60--88}.
\bibitem[{Mahapatra and Bozorgtabar(2019)}]{mahapatra2019progressive}
\bibinfo{author}{Mahapatra, D.}, \bibinfo{author}{Bozorgtabar, B.},
  \bibinfo{year}{2019}.
\newblock \bibinfo{title}{Progressive generative adversarial networks for
  medical image super resolution}.
\newblock \bibinfo{journal}{arXiv preprint arXiv:1902.02144} .
\bibitem[{Mahmood et~al.(2011)}]{mahmood2011comparison}
\bibinfo{author}{Mahmood, N.H.}, et~al., \bibinfo{year}{2011}.
\newblock \bibinfo{title}{Comparison between median, unsharp and wiener filter
  and its effect on ultrasound stomach tissue image segmentation for pyloric
  stenosis}.
\newblock \bibinfo{journal}{Inte J Appl Sci Technol} \bibinfo{volume}{1}.
\bibitem[{Moeskops et~al.(2016)}]{moeskops2016deep}
\bibinfo{author}{Moeskops, P.}, et~al., \bibinfo{year}{2016}.
\newblock \bibinfo{title}{Deep learning for multi-task medical image
  segmentation in multiple modalities}, in: \bibinfo{booktitle}{International
  Conference on Medical Image Computing and Computer-Assisted Intervention},
  \bibinfo{organization}{Springer}. pp. \bibinfo{pages}{478--486}.
\bibitem[{Nyma et~al.(2012)}]{nyma2012hybrid}
\bibinfo{author}{Nyma, A.}, et~al., \bibinfo{year}{2012}.
\newblock \bibinfo{title}{A hybrid technique for medical image segmentation}.
\newblock \bibinfo{journal}{Journal of Biomedicine and Biotechnology}
  \bibinfo{volume}{2012}.
\bibitem[{Rajaraman et~al.(2018)}]{rajaraman2018visualization}
\bibinfo{author}{Rajaraman, S.}, et~al., \bibinfo{year}{2018}.
\newblock \bibinfo{title}{Visualization and interpretation of convolutional
  neural network predictions in detecting pneumonia in pediatric chest
  radiographs}.
\newblock \bibinfo{journal}{Applied Sciences} \bibinfo{volume}{8},
  \bibinfo{pages}{1715}.
\bibitem[{Redmon et~al.(2016)Redmon, Divvala, Girshick and
  Farhadi}]{redmon2016you}
\bibinfo{author}{Redmon, J.}, \bibinfo{author}{Divvala, S.},
  \bibinfo{author}{Girshick, R.}, \bibinfo{author}{Farhadi, A.},
  \bibinfo{year}{2016}.
\newblock \bibinfo{title}{You only look once: Unified, real-time object
  detection}, in: \bibinfo{booktitle}{Proceedings of the IEEE conference on
  computer vision and pattern recognition}, pp. \bibinfo{pages}{779--788}.
\bibitem[{Standley et~al.(2020)Standley, Zamir, Chen, Guibas, Malik and
  Savarese}]{standley2020tasks}
\bibinfo{author}{Standley, T.}, \bibinfo{author}{Zamir, A.},
  \bibinfo{author}{Chen, D.}, \bibinfo{author}{Guibas, L.},
  \bibinfo{author}{Malik, J.}, \bibinfo{author}{Savarese, S.},
  \bibinfo{year}{2020}.
\newblock \bibinfo{title}{Which tasks should be learned together in multi-task
  learning?}, in: \bibinfo{booktitle}{International Conference on Machine
  Learning}, \bibinfo{organization}{PMLR}. pp. \bibinfo{pages}{9120--9132}.
\bibitem[{Teichmann et~al.(2018)}]{teichmann2018multinet}
\bibinfo{author}{Teichmann, M.}, et~al., \bibinfo{year}{2018}.
\newblock \bibinfo{title}{Multinet: Real-time joint semantic reasoning for
  autonomous driving}, in: \bibinfo{booktitle}{2018 IEEE Intelligent Vehicles
  Symposium (IV)}, \bibinfo{organization}{IEEE}. pp.
  \bibinfo{pages}{1013--1020}.
\bibitem[{Thung and Wee(2018)}]{thung2018brief}
\bibinfo{author}{Thung, K.H.}, \bibinfo{author}{Wee, C.Y.},
  \bibinfo{year}{2018}.
\newblock \bibinfo{title}{A brief review on multi-task learning}.
\newblock \bibinfo{journal}{Multimedia Tools and Applications}
  \bibinfo{volume}{77}, \bibinfo{pages}{29705--29725}.
\bibitem[{Welstead(1999)}]{welstead1999fractal}
\bibinfo{author}{Welstead, S.T.}, \bibinfo{year}{1999}.
\newblock \bibinfo{title}{Fractal and wavelet image compression techniques}.
\newblock \bibinfo{publisher}{SPIE Optical Engineering Press Bellingham,
  Washington}.
\bibitem[{Zamzmi et~al.(2020)}]{zamzmi2020accelerating}
\bibinfo{author}{Zamzmi, G.}, et~al., \bibinfo{year}{2020}.
\newblock \bibinfo{title}{Accelerating super-resolution and visual task
  analysis in medical images}.
\newblock \bibinfo{journal}{Applied Sciences} \bibinfo{volume}{10},
  \bibinfo{pages}{4282}.
\bibitem[{Zeiler et~al.(2010)}]{zeiler2010deconvolutional}
\bibinfo{author}{Zeiler, M.D.}, et~al., \bibinfo{year}{2010}.
\newblock \bibinfo{title}{Deconvolutional networks}, in:
  \bibinfo{booktitle}{2010 IEEE Computer Society Conference on computer vision
  and pattern recognition}, \bibinfo{organization}{IEEE}. pp.
  \bibinfo{pages}{2528--2535}.
\bibitem[{Zhang et~al.(2014)}]{zhang2014facial}
\bibinfo{author}{Zhang, Z.}, et~al., \bibinfo{year}{2014}.
\newblock \bibinfo{title}{Facial landmark detection by deep multi-task
  learning}, in: \bibinfo{booktitle}{European conference on computer vision},
  \bibinfo{organization}{Springer}. pp. \bibinfo{pages}{94--108}.
\bibitem[{Zhou et~al.(2017)}]{zhou2017deep}
\bibinfo{author}{Zhou, S.K.}, et~al., \bibinfo{year}{2017}.
\newblock \bibinfo{title}{Deep learning for medical image analysis}.
\newblock \bibinfo{publisher}{Academic Press}.
\bibitem[{Zhou et~al.(2019)}]{zhou2019models}
\bibinfo{author}{Zhou, Z.}, et~al., \bibinfo{year}{2019}.
\newblock \bibinfo{title}{Models genesis: Generic autodidactic models for 3d
  medical image analysis}, in: \bibinfo{booktitle}{International Conference on
  Medical Image Computing and Computer-Assisted Intervention},
  \bibinfo{organization}{Springer}. pp. \bibinfo{pages}{384--393}.
\bibitem[{Zou et~al.(2018)}]{zou2018df}
\bibinfo{author}{Zou, Y.}, et~al., \bibinfo{year}{2018}.
\newblock \bibinfo{title}{Df-net: Unsupervised joint learning of depth and flow
  using cross-task consistency}, in: \bibinfo{booktitle}{Proceedings of the
  European conference on computer vision (ECCV)}, pp. \bibinfo{pages}{36--53}.

\end{thebibliography}





\end{document}